\documentclass[sigconf]{acmart}
\usepackage[utf8x]{inputenc}
\usepackage[ruled,vlined]{algorithm2e}
\usepackage{subfig}
\usepackage{textcomp}
\usepackage{appendix}

\AtBeginDocument{%
  \providecommand\BibTeX{{%
    \normalfont B\kern-0.5em{\scshape i\kern-0.25em b}\kern-0.8em\TeX}}}

\copyrightyear{2021}
\acmYear{2021}
\setcopyright{rightsretained}
\acmConference[KDD '21]{Proceedings of the 27th ACM SIGKDD Conference on Knowledge Discovery and Data Mining}{August 14--18, 2021}{Virtual Event, Singapore}
\acmBooktitle{Proceedings of the 27th ACM SIGKDD Conference on Knowledge Discovery and Data Mining (KDD '21), August 14--18, 2021, Virtual Event, Singapore}\acmDOI{10.1145/3447548.3467162}
\acmISBN{978-1-4503-8332-5/21/08}

\begin{document}
\fancyhead{}
\title{Dynamic Language Models for Continuously Evolving Content}


\author{Spurthi Amba Hombaiah \quad Tao Chen \quad Mingyang Zhang \quad Michael Bendersky \quad Marc Najork }
\affiliation{
  \institution{Google Research}
  \city{Mountain View}
  \country{USA}
}
\email{{spurthiah, taochen, mingyang, bemike, najork}@google.com}

\begin{abstract}
The content on the web is in a constant state of flux. New entities, issues, and ideas continuously emerge, while the semantics of the existing conversation topics gradually shift. In recent years, pre-trained language models like BERT greatly improved the state-of-the-art for a large spectrum of content understanding tasks. Therefore, in this paper, we aim to study how these language models can be adapted to better handle continuously evolving web content.

In our study, we first analyze the evolution of 2013 -- 2019 Twitter data, and unequivocally confirm that a BERT model trained on past tweets would heavily deteriorate when directly applied to data from later years. Then, we investigate two possible sources of the deterioration: the semantic shift of existing tokens and the sub-optimal or failed understanding of new tokens. To this end, we both explore two different vocabulary composition methods, as well as propose three sampling methods which help in efficient incremental training for BERT-like models. Compared to a new model trained from scratch offline, our incremental training (a) reduces the training costs, (b) achieves better performance on evolving content, and (c) is suitable for online deployment. The superiority of our methods is validated using two downstream tasks. We demonstrate significant improvements when incrementally evolving the model from a particular base year, on the task of Country Hashtag Prediction, as well as on the OffensEval 2019 task. 
\end{abstract}



\maketitle

\section{Introduction}
\label{sec:introduction}

Our world is changing, and so are our languages~\cite{Aitchison2001,Kirby2007}. New entities, issues, and words are emerging rapidly. This is reflected in periodic entry additions to online dictionaries. For instance, during the Covid-19 pandemic, new words like ``Covid'' and ``Zoom'' have been added to the Oxford English Dictionary (OED)\footnote{\url{https://public.oed.com/updates/new-words-list-july-2020/}}. In addition, the usage and context of the existing words is constantly evolving to better describe our times and customs. For instance, ``flattening the curve'', which was previously an esoteric scientific term, recently became a commonplace phrase with its own sub-entry in the OED. This continuous language evolution is even more evident on the web and in social media content.

Prior works show that new words and semantic evolution pose a crucial challenge in many NLP tasks, leading to a significant performance drop for word embedding based models ({\it eg}, word2vec~\cite{Mikolov2013})
\\~\cite{Pinter2017,Kutuzov2018}.
In recent years, pre-trained transformer based language models like BERT~\cite{Devlin2019} greatly improved the state-of-the-art for a large spectrum of NLP tasks, but the study of their capability to handle dynamic content has been limited.
One relevant study by Lazaridou et al.~\cite{Lazaridou2020} shows that Transformer-XL~\cite{Dai2019}, a left-to-right language model trained on current data, still performs poorly on future instances for news and scientific articles.
A natural question is, can a bidirectional language model like BERT be successfully adapted to continuously evolving content?

\begin{figure*}[th]
    \centering
    {\includegraphics[width=0.82\textwidth]{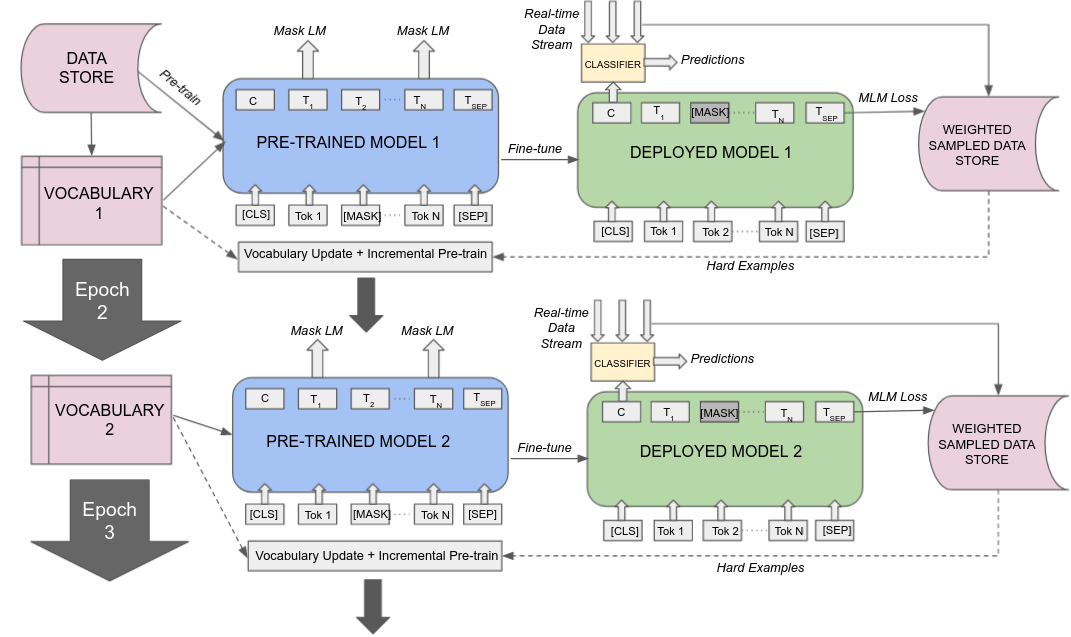}}
    \caption{Overview of System Architecture for Incremental Training of a Production Model.}
    \label{fig:system_architecture}
\end{figure*}

To answer this question, we first analyze the evolution of 2013 -- 2019 Twitter data, and unequivocally confirm that a BERT model trained on past tweets would heavily deteriorate when directly applied to data from later years. We further investigate the two possible causes for deterioration, namely, new tokens and semantic shift of existing tokens. We show (see Figure~\ref{fig:vocabulary_shift}) that there is a huge vocabulary shift over the years, {\it eg}, the most frequent words for 2014 and 2019 change by 18.31\% and 37.49\%, respectively, compared to 2013, and the most 
frequent wordpieces~\cite{Wu2016} (subwords used by BERT)
shift by roughly the same extent (see Figure~\ref{fig:vocabulary_shift_analysis}). Given this churn, wordpiece representations are likely to be sub-optimal with new data, leading to a decrease in the effectiveness of the learned representations.

Therefore, we propose to dynamically update the wordpiece vocabulary, by adding emerging wordpieces and removing stale ones, aiming at keeping the vocabulary up-to-date, while maintaining its constant size for ensuring efficient model parameterization. In addition, we examine two different vocabulary composition methods for Twitter hashtags: (a) feeding each hashtag after stripping ``\#'' to the WordPiece tokenizer and (b) retaining whole popular hashtags as tokens in the wordpiece vocabulary, as they may capture some of the current zeitgeist semantics. We notice that keeping popular whole hashtags in the vocabulary could bring over 25\% gain across different metrics for hashtag sensitive tasks. 

To examine the semantic shift, we select a few country hashtags as a case study. By comparing their top co-occurring hashtags and words, we show that the semantics of the country hashtags shift over the years. We, thus, propose to incrementally pre-train BERT with new data as it appears, so that the model can adapt to the language evolution. 
However, simply using all new data can be very costly, as training BERT is computationally expensive~\cite{Sharir2020}. To reduce the amount of the required training data, we propose three effective sampling approaches to iteratively mine representative examples that contain new tokens, or tokens which potentially exhibit large semantic shifts, for incremental learning. 

Our incremental learning reduces the training cost by 76.9\% compared to training an entirely new model, while also achieving better prevention of model deterioration as new content emerges. We evaluate the model performance on two downstream tasks on a large Twitter dataset -- Country Hashtag Prediction and offensive tweet prediction (OffensEval 2019 task~\cite{Zampieri2019}). We demonstrate significant improvements for our incremental training methods which use effective sampling over baselines in these evaluations.

To deploy our model in production, we first generate model vocabulary using a particular year's data, pre-train the model, and fine-tune it using task data. Figure~\ref{fig:system_architecture} gives an overview of our proposed architecture. We continuously monitor the MLM loss on real-time data stream and on detecting performance deterioration for the current model, we draw hard examples from a weighted data store using an effective sampling strategy described in Section ~\ref{sec:token-mlm-loss}. We update the model vocabulary and incrementally train the model using the hard examples. The model is further fine-tuned and deployed. In this way, the entire life-cycle of dynamic model updates (vocabulary update, pre-training, and fine-tuning) can occur while continuously serving live traffic.

To summarize, the main contributions of this work are as follows:
\begin{itemize}
    \item To the best of our knowledge, we are the first to study dynamic BERT modeling for continuously evolving content.
    \item We propose a simple yet effective method to dynamically update BERT model vocabulary. 
    \item We observe that keeping popular whole hashtags in model vocabulary can benefit certain tasks and validate our dynamic BERT modeling technique based on two different model vocabulary compositions.
    \item We propose three different sampling methods for more efficient incremental BERT training based on hard example mining. 
    \item One of our proposed methods can also be used to determine when incremental training should be triggered in real-world applications. 
\end{itemize}

\vspace*{-2pt}
\section{Related Work}
\label{sec:related_work}
As language evolves, new words are emerging and the semantics of existing words are drifting~\cite{Aitchison2001,Kirby2007}. In this section, we first discuss how the prior work addresses these two challenges in language modeling, and then summarize the existing work on incremental learning (which is applied in our work in the context of dynamic language modeling).

\vspace*{-2pt}
\subsection{Handling New Words}

New words that are out of vocabulary (OOV) pose great challenges to many NLP tasks~\cite{Pinter2017}.
The model performance could be significantly hurt by a high OOV rate, especially for morphologically rich languages and domains with dynamic vocabularies ({\it eg}, social media)~\cite{Kim2016}. Simply designing a language model with overly large vocabularies cannot completely resolve the OOV issue, as new words are always emerging, while also being parametrically expensive~\cite{Mi2016,Sennrich2016}.

In language modeling, several approaches have been proposed to address this issue. As the embeddings of new words do not exist in the training data, one line of work replaces all new words by a special token ({\it eg}, "UNK") 
with shared embeddings~\cite{Hermann2015} or assigns unique random embeddings to each new word~\cite{Dhingra2017}. In a separate line of studies, researchers break-down a word to more fine-grained units, including characters~\cite{Zhang2015,Ling2015,Kim2016,Pinter2017,Al-Rfou2019}, character-based n-grams~\cite{Wieting2016,Bojanowski2017,Takase2019}, and subwords ({\it eg}, wordpiece~\cite{Wu2016} and byte-pair-encodings~\cite{Sennrich2016,Kudo2018}). This could reduce OOV rate since these fine-grained units are less likely to be unseen at the training stage. From modeling aspect, these prior works leverage the morphological structure for learning embeddings, and often adopt a pooling layer ({\it eg}, CNN, LSTM) to combine the embeddings of fine-grained units to construct the word embeddings. One limitation of this direction is that some words can not be inferred from their subunits ({\it eg}, a person's name or a Twitter hashtag).

The third line of research attempts to explicitly generate OOV word embeddings ``on the fly'' from context such as the definitions of the OOV word in a dictionary~\cite{Bahdanau2018} and 
example sentences that contain the OOV word~\cite{Lazaridou2017,Herbelot2017,Khodak2018,Hu2019}. Most works adopt a simple pooling approach, {\it eg}, summation~\cite{Herbelot2017}, 
mean pooling~\cite{Bahdanau2018,Khodak2018}
to aggregate the embeddings of the contextual words as the OOV word embeddings, while Hu et al.~\cite{Hu2019} propose an attention-based hierarchical context encoder to encode and aggregate both context and subword information. In a multilingual setting, 
Wang et al.~\cite{Wang2019} adopt joint and mixture mapping methods from pre-trained embeddings of low resource languages to that of English at subword level to address this. 

In our work, we adopt the Transformer-based language model BERT~\cite{Devlin2019}, which uses wordpieces as the basic units. Though the prior work shows that subword representation is a useful strategy for dealing with new words, we show that there is still a significant model downgrade for ever evolving content like Twitter. We, thus, propose to dynamically update the vocabularies by swapping the stale tokens with the popular emerging ones.

\vspace*{-\baselineskip}
\subsection{Semantic Shift over Time}
The semantics of existing words keep evolving. Kutuzov et al.~\cite{Kutuzov2018} conduct a comprehensive review on this topic, and we only briefly discuss the most relevant works here.
As case studies, early works choose a few words to discuss their semantic drifts over widely different time periods~\cite{Ullmann1962, Blank1999, Traugott2001}. The more recent works aim to automatically detect word semantic changes where the semantics of words are in distributional representation ({\it eg}, word-context matrix)~\cite{Sagi2009, Gulordava2011, Cook2010} or, recently more popular, distributed representation ({\it ie}, word embeddings)~\cite{Hamilton2016}. These works usually train different word representation models with documents from different time slices, and then compare the word semantic representations over time using cosine distance to quantify the semantic change. 

As each word embedding model is trained separately, the learned embeddings across time may not be placed in the same latent space~\cite{Kutuzov2018}. Several approaches have been proposed to resolve this alignment issue as a second step~\cite{Zhang2015b, Kulkarni2015, Hamilton2016}. For instance, Kulkarni et al.~\cite{Kulkarni2015} use a linear transformation that preserves general vector space structure to align learned embeddings across time-periods and Hamilton et al.~\cite{Hamilton2016} use orthogonal Procrustes to perform embedding alignments while preserving cosine similarities. 

Other works attempt to simultaneously learn time-aware embeddings over all time periods and resolve the alignment problem~\cite{Yao2018,Bamler2017,Rudolph2016}. Yao et al.~\cite{Yao2018} propose to enforce alignment through regularization,  Bamler et al.~\cite{Bamler2017} develop a dynamic skip-gram model that combines a Bayesian version of the skip-gram model~\cite{Barkan2017} with a latent time series, and Rudolph et al.~\cite{Rudolph2018} propose dynamic embeddings built on exponential family embeddings to capture sequential changes in the representation of the data.

Though there are plenty of prior works, most of them are based on non-contextualized embeddings and limited work has been done for Transformer-based language models. The most relevant work is by \cite{Lazaridou2020} who demonstrate that Transformer-XL~\cite{Dai2019} (a left-to-right autoregressive language model) handles semantic shifts poorly in news and scientific domains, and highlight the importance of adapting language models to continuous stream of new information.
In this work, we aim to bridge this gap and propose to detect and adapt to semantic drift using BERT in a training framework based on the incremental learning research.

\vspace*{-5pt}
\subsection{Incremental Learning}
\label{sec:incremental-training}

Incremental learning is a family of machine learning methods that use continuous input data ({\it eg}, data streams) to expand the capability of an existing model, such as gradually increasing the number of classes for a classifier. One challenge that incremental learning faces is \textit{catastrophic forgetting}, namely a dramatic performance decrease on the old classes when training data with new classes is being added incrementally~\cite{Castro2018}. This is even more evident for the deep learning models~\cite{Shmelkov2017,Rebuffi2017}. Training a model from scratch with both old and new data seems to remedy this issue but is expensive in terms of computational resources as well as carbon emissions~\cite{Strubell2019}.
To mitigate this, one line of work proposes to select a representative memory from the old data and then, incrementally train the model with both memory and the new data~\cite{Castro2018}. Other works utilize distillation loss~\cite{Jung2016} aiming at retaining the knowledge from old classes, and combining this with the standard cross-entropy loss to learn to classify the classes~\cite{Li2017,Castro2018}. 

In our work, we adopt an incremental learning framework to build the dynamic BERT model based on continuously evolving content where the emerging vocabulary entries can be considered as new classes.
Different from typical incremental learning, the semantics of existing words ({\it ie}, old classes) may also change over time. As such, we propose to intentionally update/forget the information of old classes that have an obvious semantic drift. This work is also different from the BERTweet model described in ~\cite{Nguyen2020} which is pre-trained with data over several years (2012 -- 2019), whereas our models are incrementally trained to keep their performance on evolving content, from base models pre-trained with a particular year's data.

\section{Dynamic Language Modeling}
\label{sec:dynamic_modeling}

Our language is continuously evolving, especially for the content on the web. Can a language model like BERT that is pre-trained on a large dataset adapt well to the evolving content? To answer this, we use a large public Twitter corpus crawled from 2013 -- 2019 for preliminary experiments and analysis.
We pre-train year-based Twitter BERT models on the Masked Language Modeling (MLM) task, using the tweets from a particular year. All of them are base models (12 layers) and initialized using the public BERT pre-trained on Wikipedia and books~\cite{Devlin2019}. For model evaluation, we use two downstream tasks, Country Hashtag Prediction (predicting the associated country hashtag for a tweet from 16 pre-selected country hashtags)  and OffensEval 2019~\cite{Zampieri2019} (a shared task from SemEval 2019 to predict if a tweet is offensive or not). The data for Country Hashtag Prediction is curated from the 2014 and 2017 tweets in our dataset, while OffensEval is using tweets posted in 2019. The dataset and experiments are detailed in Section~\ref{sec:experiment}.

Our results unequivocally show that BERT pre-trained on past tweets heavily deteriorates when directly applied to data from later years. Take the results of 2017 Country Hashtag Prediction as an example (Table~\ref{table:2017_hashtag_base_results}). The 2016 model achieves $0.418$ in Micro-F1, $0.265$ in Macro-F1, and $0.411$ in Accuracy, which is significantly worse than the 2017 model ($0.561$ in Micro-F1, $0.493$ in Macro-F1, and $0.550$ in Accuracy). This suggests the necessity to keep the model informed of the evolving content. To gain more insights, we investigate two possible causes for the performance deterioration: (a) vocabulary shift and (b) semantic shift of existing words, and propose dynamic modeling solutions to address these two challenges. 

\begin{table}[th]
\small
\caption{Results on 2017 Country Hashtag Prediction.}
\label{table:2017_hashtag_base_results}
\begin{tabular}{|c|c|c|c|}
\hline
\textbf{Model}   & \textbf{Micro-F1} & \textbf{Macro-F1} & \textbf{Accuracy} \\ \hline
Base Model 2016 & $0.418 \pm 0.003$ & $0.265 \pm 0.002$ & $0.411 \pm 0.003$ \\
\hline
Base Model 2017 & $0.561 \pm 0.003$ & $0.493 \pm 0.003$ & $0.550 \pm 0.003$ \\
\hline
\end{tabular}
\end{table}

\vspace*{-\baselineskip}
\begin{figure}[th]
    \centering
    \includegraphics[width=0.3\textwidth]{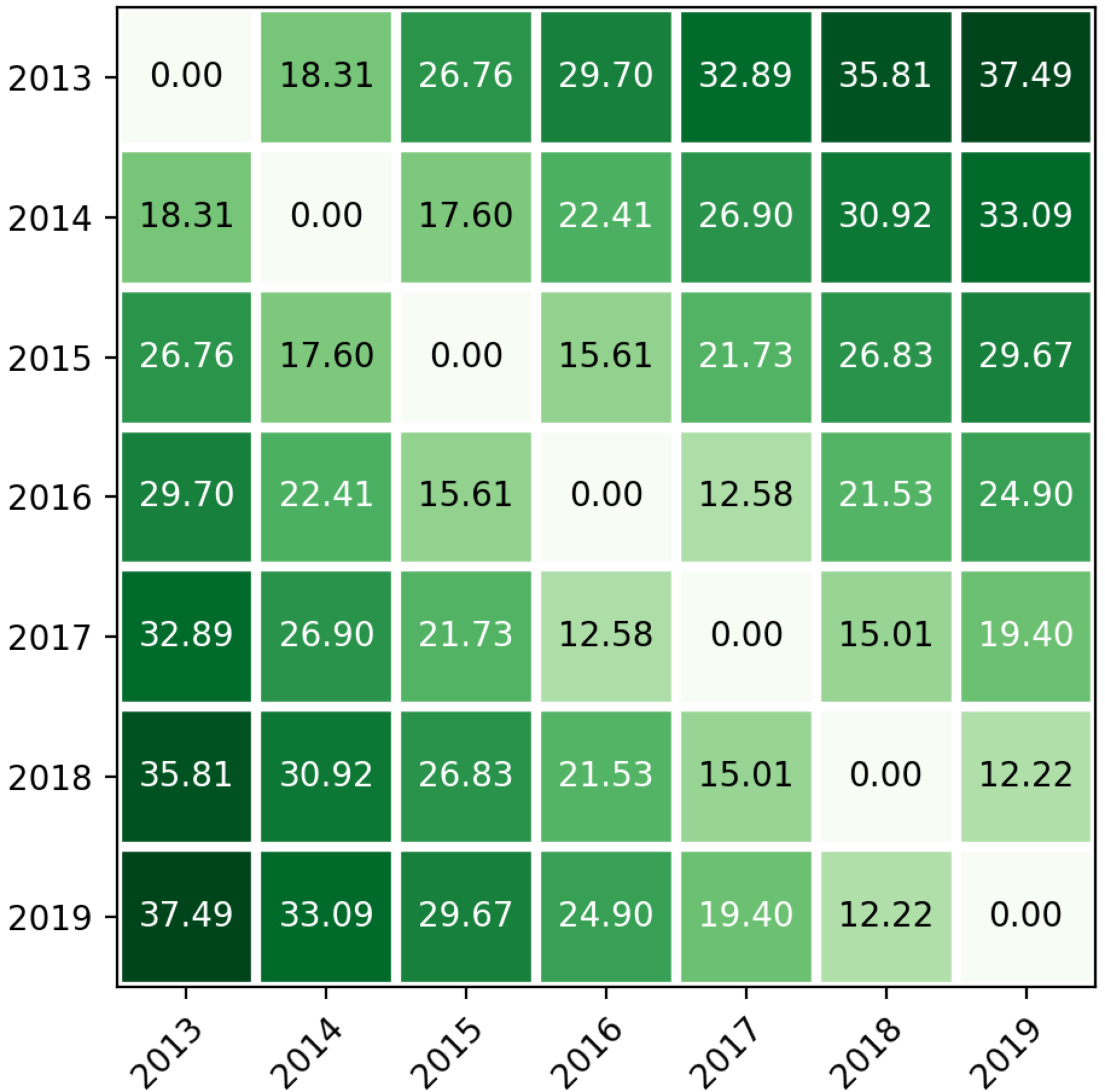}
    \caption{Vocabulary shift (\%) for natural words using the top 40k tokens. Corresponding figures for wordpieces and hashtags can be found in Appendix~\ref{subsec:vocab_shift_analysis}.}
    \label{fig:vocabulary_shift}
\end{figure}

\vspace*{-\baselineskip}
\subsection{Vocabulary Shift}
\label{sec:vocab_shift}
Vocabulary is the foundation for language models. Vocabulary can consist of natural words and more fine-grained units like subwords ({\it eg}, wordpieces), character-based n-grams, or even single characters. Out of vocabulary (OOV) tokens pose great challenge to language models as their embeddings do not exist in the model training~\cite{Pinter2017,Kim2016}. To deal with this, a common practice is to map the OOV tokens to a special ``UNK'' token such that all OOV tokens share the same embeddings. Obviously, shared embeddings lose specificity and are not informative. Prior works~\cite{Zhang2015,Kim2016,Pinter2017,Wieting2016,Wu2016} show that fine-grained units are effective in reducing OOV rate as a new/unseen word could still be broken down into existing tokens in the vocabulary. A natural question is, can wordpieces that are adopted by BERT adapt well to the new words on Twitter? To this end, we conduct wordpiece vocabulary shift analysis. Moreover, we perform similar analysis for natural words and hashtags. We first describe the three token variants in detail:

\begin{itemize}
	\item \emph{Natural Words} These are innate vocabulary tokens commonly used by humans. Their change directly reflects changes in the general language.
	\item \emph{Subword Segments} WordPiece~\cite{Devlin2019} and SentencePiece~\cite{Kudo2018} are arguably the two most popular methods for machine language tokenization. They both break-down natural words into subwords, and they attest the fact that subword segmentations are not only more effectively utilized by machines, but can also reduce the size of the vocabulary. In this paper, we adopt the WordPiece method but our discussions can be applied to any tokenization method. 
	\item \emph{Hashtags} These are special tokens that start with a ``\#'' symbol, widely used on social media platforms like Twitter, Facebook, and Instagram. Compared to natural words, hashtags have a higher change rate. A hashtag can be a label of a message, or can be directly a part of the message content. Hashtags are extremely important for dynamic content modeling, since they often indicate the key topics of the social media post. 
\end{itemize}

Based on our 2013 -- 2019 Twitter dataset, we create the top 40K vocabulary for natural words, wordpieces, and hashtags in each year. 
All tokens are lowercased in pre-processing. For wordpieces, the WordPiece tokenizer is applied to each year's tweets separately. 
We then compare these vocabularies and plot their shift rates for natural words in Figure~\ref{fig:vocabulary_shift}, and for wordpieces and hashtags in Figure~\ref{fig:vocabulary_shift_analysis} in Appendix~\ref{subsec:vocab_shift_analysis}.
The shift is defined as:
\begin{center}
${1.0 - \frac{ |Vocab\_1 \bigcap  Vocab\_2|}{|Vocab\_1 \bigcup  Vocab\_2|}}$\\
\end{center}
We see that all three types of tokens have huge vocabulary shifts. Among them, hashtags exhibit the largest year-over-year shifts: 2014 and 2019 change by 58.75\% and 78.31\%, respectively, compared to 2013. Since hashtags are the topic indicators for the posts, these huge shifts also validate that the content on Twitter is drastically evolving. Natural words change by 18.31\% and 37.49\% for 2014 and 2019, respectively, compared to 2013. Wordpieces follow similar trends, changing 19.63\% and 38.47\% in 2014 and 2019, respectively, compared to 2013. 

Note that our analysis is based on the top 40K tokens. It is likely that using a larger vocabulary may reduce the year-to-year shifts/OOV rates. However, the memory limitation and computational cost prohibit extremely large vocabulary for mainstream pre-trained language models. Most models are only able to keep tens of thousands of tokens in the model vocabulary. For instance, the original BERT uses 30K wordpieces~\cite{Devlin2019}. Using large vocabularies would make models parametrically expensive and render them infeasible for real-world applications/deployment.

\subsection{Sub-optimal Tokenization for New Words}
\label{sec:sub-optimiazed-tokenization}
Though the year-to-year vocabulary discrepancies are huge, we observe that the actual wordpiece OOV rate is low when applying a model to data from later years. For instance, with the wordpiece vocabulary curated from 2013 tweets, the OOV rate for 2014 data is 0.54\%. The reason is that the WordPiece tokenizer could still decompose a new/unseen word from later years into known subwords or even characters. However, this does not necessarily guarantee that the semantics of the new word is well preserved.
For instance, the words ``griezmann'' and ``\#unesco'' from 2014 data are tokenized into the wordpieces \{``gr'', ``\#\#ie'', ``\#\#zman'', ``\#\#n''\} and \{``\#un'', ``\#\#es'', ``\#\#co''\}, respectively, using the 2013 vocabulary. It is difficult for BERT to capture the correct semantics from these wordpieces. 

To further investigate this, we replace the wordpiece vocabulary of a 2017 model with the vocabulary from 2013 data, and retrain the model on 2017 data. For the 2017 Country Hashtag Prediction task, we observe that using an out-dated vocabulary decreases the model performance by 6.57\% (in terms of Micro-F1), relative to using the vocabulary from the same year. This confirms that subword representation like wordpieces is not an optimal solution to handle new words in rapidly evolving content. 

\vspace*{-3pt}
\subsection{Vocabulary Composition for Hashtags}
\label{sec:hashtag_compsition}

As hashtags often mark the topics in the posts, we believe that understanding hashtags is key to the language model quality. 
Hashtags can consist of a single word ({\it eg}, ``\#amazon''), multiple words ({\it eg}, ``\#wordcup2014''), or some characters indicating an abbreviation ({\it eg}, ``\#nfl'').
There are two straightforward approaches to incorporate hashtags into the modeling. One is to strip the ``\#'' and treat hashtags as normal natural words, feeding them to the WordPiece tokenizer. 
It is very likely that many hashtags, especially those that have multiple words, are segmented into subwords or even characters. The strong topical information may be lost due to the segmentation. Hashtag ``\#ItsComingHome'', which means winning the Football World Cup is such an example. The WordPiece tokenizer decomposes it into three wordpieces ``Its'', ``Coming'', and ``Home'', which, however, poses difficulty for the model to relate these three wordpieces to their original meaning. To alleviate this, the second method to model hashtags is to include popular hashtags (with ``\#') as intact tokens in the wordpiece vocabulary and only tokenize rarer ones as ordinary words. 

We compare the two hashtag vocabulary composition approaches on the aforementioned downstream tasks -- Country Hashtag Prediction and OffensEval. From Table~\ref{table:hashtag-analysis-hashtag-prediction-task}, we see that including hashtags in the vocabulary largely boosts the model performance for the 2017 Country Hashtag Prediction task (using a model trained from scratch with 2017 data), improving the Micro-F1 from 0.314 to 0.561. On the other hand, for OffensEval (using a model trained from scratch with 2019 data), including hashtags does not bring any gains and slightly hurts the model performance as shown in Table ~\ref{table:hashtag-analysis-offenseval}.

We attribute these different effects to the nature of the two tasks. For the Country Hashtag Prediction task, the model needs to understand the topics covered in the post well, and then make a prediction about the associated country. Hashtag tokens carry more contextual information than ordinary words. For instance, a country hashtag could carry semantics of events associated with this country and, would not just be limited to a regular country name that indicates a geographic location. Therefore, differentiating hashtags and regular words in the vocabulary is beneficial for this task. On the other hand, for the task of OffensEval, the dataset itself does not contain many hashtags, and most hashtags are not informative to determine whether a tweet is offensive or not. As such, including intact hashtags in the vocabulary is not beneficial.

Based on these results, in the remainder of the paper, for the task of Country Hashtag Prediction, we will include popular whole hashtags in model vocabulary; for the task of OffensEval, we will break-down all hashtags into wordpieces, after stripping ``\#''.

\begin{table}[ht]
\begin{center}
\caption{Performance using different Hashtag Vocabulary Composition for 2017 Country Hashtag Prediction.}
\label{table:hashtag-analysis-hashtag-prediction-task}
\small
\begin{tabular}{|p{3.167cm}|p{1.36cm}|p{1.36cm}|p{1.36cm}|}
\hline
\textbf{Vocabulary Composition}  & \textbf{Micro-F1} & \textbf{Macro-F1} & \textbf{Accuracy} \\
\hline
Include Whole Hashtags & $0.561 \pm 0.003$ & $0.493 \pm 0.003$ & $0.550 \pm 0.003$ \\
\hline
Break-down Hashtags & $0.314 \pm 0.002$ & $0.156 \pm 0.001$ & $0.308 \pm 0.003$ \\
\hline
\end{tabular}
\end{center}
\end{table}

\vspace*{-\baselineskip}
\begin{table}[ht]
\begin{center}
\caption{Performance using different Hashtag Vocabulary Composition for OffensEval 2019.}
\label{table:hashtag-analysis-offenseval}
\small
\begin{tabular}{|c|c|c|}
\hline
\textbf{Vocabulary Composition} & \textbf{F1} & \textbf{AUC-ROC} \\
\hline
Include Whole Hashtags & $0.491 \pm 0.015$ & $0.567 \pm 0.015$ \\
\hline
Break-down Hashtags & $0.506 \pm 0.010$ & $0.636 \pm 0.011$ \\
\hline
\end{tabular}
\end{center}
\end{table}

\vspace*{-3pt}
\subsection{Dynamic Updates to Model Vocabulary}
\label{sec:dynamically-update-model-vocabulary}

As we discussed in Section~\ref{sec:vocab_shift}, wordpieces are not effective in handling rapidly evolving content that exhibits large vocabulary shifts. Instead of leveraging a static vocabulary, we argue that it is vital to dynamically update model vocabulary to reflect the evolving content. To this end, we propose a simple yet highly effective algorithm to add the most frequent new wordpieces and remove the outdated ones ({\it ie}, least likely to occur in the new wordpieces) from the vocabulary. We detail this approach in Algorithm~\ref{alg:model_vocab_update} in Appendix~\ref{subsec:vocab_update_algo}. For hashtag sensitive tasks like Country Hashtag Prediction, we also add/remove popular/unpopular whole hashtags in the vocabulary. Our goal is to keep the vocabulary up-to-date, while maintaining its constant size for ensuring efficient model parameterization.

After replacing the outdated tokens with new ones, we continuously train the model with data sampled from the new timestamp. We will detail the training strategies in Section~\ref{sec:effective_sampling_for_incremental_training}. Our later experiments show that this vocabulary updating approach is very beneficial for model performance (detailed in Section~\ref{sec:experiment}).

\subsection{Token Semantic Shift}
\label{sec:token-semantic-shift}
Aside from emerging words, it is well known that the semantics of existing words keep evolving~\cite{Kutuzov2018, Ullmann1962, Blank1999, Traugott2001}. To measure the semantic shift, one intuitive way is to compare their embeddings learned in different years. However, since each year's BERT model was trained separately and their semantic space may not be well aligned, direct comparisons may not be meaningful. Instead, we turn to the contextual words as a proxy of semantic representation. 

We use country hashtags in our Country Hashtag Prediction task as a case study. We pick 1,000 most frequently co-occurring words for the hashtags from 2014 and 2017 dataset, to confirm that the semantics are shifting significantly. Taking the three country hashtags  ``\#china'',  ``\#uk'', and ``\#usa'' as examples, we compute the rates of shift in top contextual words for these hashtags as 44.07\%, 45.80\%, and 65.59\%, respectively.
These significant shifts can be explained by the widely varying topics seen for 2014 and 2017 for the respective countries.
For instance, for the hashtag ``\#usa'', many of the top topics ({\it eg}, ``\#worldcup'', ``ronaldo'') for 2014 revolve around the Football World Cup, whereas in 2017, several top topics ({\it eg}, ``\#maga'', ``\#theresistance'') concern important developments in the US politics.
Table~\ref{table:topics-with-different-country-hashtags} in Appendix~\ref{subsec:topics-with-different-country-hashtags} further shows five of the top co-occurring words for these hashtags that are representative of the topics and events. As with the prior work, we propose to continuously train the model with updated data to handle the semantic shift which is detailed in the following section.
\section{Effective Sampling for Incremental Training}
\label{sec:effective_sampling_for_incremental_training}

For our proposed approach, we aim to dynamically update the vocabulary -- adding new tokens and removing obsolete ones --  and adapt the semantics of the tokens to reflect the evolving content. In addition to these vocabulary shifts, new web and social content is being continuously created en masse, {\it eg}, on an average, 500 million tweets are posted everyday and 200 billion tweets are created per year\footnote{\url{https://www.dsayce.com/social-media/tweets-day}}. These motivate us to adopt an incremental learning framework to build our dynamic BERT model. 
As with typical incremental training~\cite{Castro2018, Jung2016}, we need to learn new knowledge ({\it eg}, the semantics for new words) while retaining the model's existing knowledge ({\it eg}, keeping the meanings of words that do not have a semantic shift). In our case, however, we also need to intentionally update model's existing knowledge on those tokens which have a semantic shift.

One key component of incremental training is to select proper data to further train the model~\cite{Castro2018}. Naively, we could use all tweets from the latest year to continuously train the model built previously. However, training models like BERT are known to be computationally expensive, particularly with a large dataset such as an entire year of tweets. To reduce the training cost and make the incremental training feasible, one simple approach is to randomly sample some sizable data from the new year's tweets as the training dataset. However, a random sample may not fully capture the evolution of the language. 

We, thus, propose three sampling approaches to mine representative tweets that contain evolving content ({\it eg}, new tokens or tokens that are likely to have undergone semantic shift), which is in the spirit of active learning.
Our intuition is that new instances tend to contain evolving content if the current model performs poorly on them, or their embeddings have changed dramatically since the training of the last model. We detail the three approaches below. All three approaches run iteratively to detect representative examples and keep improving the model.
In addition, we would like to highlight that the application of our proposed methods is not limited to continuously evolving content, but can also be applied to any scenario in which knowledge shift happens.

\subsection{Token Embedding Shift Method}

We leverage the change of a token's embedding as a signal for evolving language. In each iteration, we compute the cosine distance between a token's embedding from the updated model and its preceding version.
For the first iteration of training, we compare the incremental model vocabulary with the base model's vocabulary to identify new tokens. We give higher weights to tweets containing new tokens when sampling. For successive iterations, we identify top $X$ tokens which exhibit the largest shift in their embeddings between the current model and its preceding version, where $X$ is domain dependent ({\it ie}, how fast the vocabulary evolves between successive time-periods).
When sampling, we assign large weights to the tweets that contain tokens with large embedding shift. In addition, we observe that tokens in a short tweet tend to have a larger embedding shift. Therefore, we linearly combine embedding cosine distance and normalized tweet length as the sampling weight.

Algorithm~\ref{alg:weighted_random_sampling} in Appendix~\ref{subsec:incremental_training_algo} details this iterative approach. 
In the first iteration, we randomly sample some tokens if the vocabulary does not change; otherwise, we pick tokens that are newly added to the vocabulary. In the later iterations, we use tokens' shift in embeddings to perform a weighted random sampling and then, continuously train the model. We repeat this process for $n$ iterations, where $n$ is a tunable parameter. 
\vspace{-6pt}
\subsection{Sentence Embedding Shift Method}

Similar to the token embedding shift method, we measure the embedding shift via cosine distance for a sentence ({\it ie}, a tweet) between the updated model and its previous version. Following the convention, we consider the [CLS] token embedding as the sentence embedding. Again, longer sentences are assigned a larger weight because short sentences tend to have larger embedding variances. We use the combination of embedding cosine distance and tweet length to perform weighted random sampling, and iteratively update the model for $n$ iterations (detailed in Algorithm~\ref{alg:weighted_random_sampling} in Appendix~\ref{subsec:incremental_training_algo}).

\vspace{-5pt}
\subsection{Token MLM Loss Method}
\label{sec:token-mlm-loss}

Token Masked Language Modeling (MLM) loss is a pre-training loss proposed by BERT. It measures whether a model can successfully predict a token when the token is masked out from the model's input. 
Different from its original form in BERT pre-training, we can apply it to either a pre-trained or a fine-tuned model to identify tweets with token semantic shift. Here, we modify the task definition to fit our use-case. We don't mask out any tokens from the model input. Instead, we take the last layer of the pre-trained BERT, directly mask out tokens from that layer, and then use the surrounding tokens from the same layer to predict the masked tokens. The benefit of doing this is as follows: when a model (fine-tuned on some task(s)) is being served online, we don't need to change either the model's input or output to calculate the new MLM loss. When the fine-tuned model is inferred, we just need to take the last layer of the pre-trained model (not that of the fine-tuned model) and compute the losses. The model's online serving quality won't be affected and the token MLM loss calculation is not only light-weight, but can also be piggy-backed to model serving. This method can also run iteratively using the proposed Algorithm~\ref{alg:weighted_random_sampling} (Appendix~\ref{subsec:incremental_training_algo}).

\paragraph{Deployed Model}
Figure~\ref{fig:system_architecture} shows the conceptual architecture of a production system based on our incremental training method.
The initial model is pre-trained using vocabulary and tweets derived from a particular ``base'' time-period. This base model is further fine-tuned with task specific data and deployed to serve real-time traffic. For incremental training, ``Token MLM Loss'' sampling strategy is used to mine representative tweets because of its strong performance and unique benefits (elaborated in Section~\ref{sec:further-discussion}).

During model serving, token MLM loss is additionally computed and stored with the data. Whenever there is a significant MLM loss increase
on the new data, a new incremental training epoch will be triggered. We draw hard examples from the new data, update the model vocabulary, and incrementally pre-train with these examples. We then fine-tune the model for the specific task, and deploy the resulting model. We continue to train new epochs as needed, to keep the model up-to-date with the evolving data stream.
\section{Experiments}
\label{sec:experiment}

In this section, we evaluate our proposed dynamic modeling and efficient incremental training strategies on rapidly evolving Twitter content. We choose to use Twitter data for our experiments as it is one of the large scale publicly available datasets. We describe the experimental settings for model pre-training, training cost savings, two downstream tasks for model evaluation, and conclude by discussing the experimental results.

\subsection{Pre-training}
We describe the data and detail its pre-processing in Appendix~\ref{subsec:data_preprocessing}.
In all our experiments, we use a 12 layer BERT and MLM loss as our pre-training objective. To simplify our discussion, we define two types of models:
\begin{itemize}
	\item \emph{Base Model} This is a fully trained model with one year's tweets. We initialize it using the original BERT-Base model
	\\~\cite{Devlin2019}.
	Its vocabulary is updated once from the original BERT-Base model vocabulary using Algorithm~\ref{alg:model_vocab_update} (Section ~\ref{sec:dynamically-update-model-vocabulary}) except that we do not remove any tokens from the original vocabulary (as its size is around 30k). In other words, base model vocabulary is the union of BERT-Base model vocabulary and optimized wordpieces (and hashtags if the vocabulary composition includes whole hashtags) from that year.
	\item \emph{Incremental Model} This is a model incrementally trained based on the previous year's base model. Its vocabulary is iteratively updated from the prior year's base model vocabulary, again using Algorithm ~\ref{alg:model_vocab_update} (Section ~\ref{sec:dynamically-update-model-vocabulary}). We initialize the embeddings for common tokens for the incremental model from the corresponding embeddings of the base model and randomly initialize the embeddings for the newly added tokens, when we start training.
\end{itemize}
Note that we opt to train incremental models using the previous year's base model only and not using the data accumulated over several past years. This simulates the effectiveness of continuously adapting a trained model in a production setting (serving online traffic).
	
For both models, we keep the vocabulary size fixed at 65K. When whole hashtags are included in the vocabulary, we reserve 15K for them and use the rest 50K for wordpieces generated by the WordPiece tokenizer.

For every year, we randomly split the sampled 50M tweets into 45M for training and 5M for evaluation. All base models are trained for 2.6M steps with 45M tweets. For incremental models, we start from a 2M step base model checkpoint from previous year, sample some new year's tweets and incrementally train the model for additional 600k steps. This strict setting to have both kinds of models trained for identical number of steps (2.6M) aims to ensure a fair comparison between their results. 

For incremental training, 
we implement two simple sampling methods as baselines to compare against our proposed sampling approaches (Section ~\ref{sec:effective_sampling_for_incremental_training}):
\begin{itemize}
    \item \emph{Uniform Random Sampling} We draw a sample uniformly at random from the new year's tweets.
    \item \emph{Weighted Random Sampling} We draw a random sample weig-\\
    hted by the number of wordpiece tokens in the tweet. Since longer tweets tend to be more informative and contain evolving content, we favor longer tweets than shorter ones in the sampling. 
    This method shows some empirical benefits in our experiments. 
\end{itemize}

Each baseline samples 24M tweets from the 45M pool to continuously train a model starting with a base model from previous year for an additional 600K steps. An incremental model is trained iteratively, {\it ie}, sampling new tweets using our proposed sampling methods and updating the model in each iteration. We empirically use three iterations, and in each iteration train the model for 200K steps with newly sampled tweets in this iteration together with all tweets sampled in the previous iterations. To be specific, we draw a sample of 10M, 8M, and 6M tweets for the first, second and, third iterations, respectively. All sampling is performed without replacement. Our incremental models will see 24M unique tweets in total (other than the 45M examples used for training the base model), which is the same amount of tweets used for baseline sampling models. We describe the hyperparameters used for training in Appendix~\ref{subsec:hyperparameters}.

\vspace*{-5pt}
\subsection{Training Cost Savings}
\label{subsec:cost_savings}
Compared to training a base model from scratch (2.6M steps), our proposed architecture for training an incremental model in Figure~\ref{fig:system_architecture} significantly reduces the training cost. Since the cost of incremental training is only 600k steps, we save 2M steps which yields a cost savings of 76.9\% relative to the base model.

\subsection{Evaluation}
\label{subsec:downstream}
As briefly described in Section~\ref{sec:dynamic_modeling}, we assess the model performance with two downstream tasks:
\begin{itemize}
    \item{Country Hashtag Prediction (2014 and 2017)}: This task aims to predict the associated country hashtag for a tweet from a pre-defined country list (detailed in Appendix~\ref{subsec:country_hashtag_prediction_task}).
    
    Note that training multiple end-to-end models for all years is resource intensive in terms of both compute and time. Hence, these two years which form a representative subset of all years (2013 -- 2019) were chosen for our experiments. 
    
    \item{OffensEval 2019}: OffensEval is one of the tasks under SemEval aimed at identifying if tweets are offensive (detailed in Appendix~\ref{subsec:offenseval_task}).
\end{itemize}

We expect Country Hashtag Prediction to be more sensitive to topical content like hashtags and semantically shifted words, while the OffensEval task (similar to other NLP tasks like sentiment analysis) is less so. We would like to evaluate our proposed architecture on both types of tasks. 

For all the downstream tasks, we fine-tune pre-trained models for 300K steps. As Country Hashtag Prediction is a multi-class classification task, we report micro-F1, macro-F1, and accuracy scores for all the models on the test set. Since OffensEval is a binary classification task, we report F1 score and AUC-ROC.

\vspace*{-\baselineskip}
\begin{table}[ht]
\begin{center}
\caption{Results for 2014 Country Hashtag Prediction.}
\label{table:country-hashtag-prediction-2014}
\small
\begin{tabular}{|c|c|c|c|}
\hline
\textbf{\hfil{Model}} & \textbf{\hfil{Micro-F1}} & \textbf{\hfil{Macro-F1}} & \textbf{\hfil{Accuracy}} \\
\hline
Base Model 2013 & $0.124 \pm 0.002$ & $0.014 \pm 0.000$ & $0.121 \pm 0.001$ \\
\hline
Base Model 2014 & $0.467 \pm 0.002$	& $0.357 \pm 0.002$	& $0.456 \pm 0.002$ \\
\hline
\hline
Uniform Random & $0.583 \pm 0.003$ & $0.495 \pm 0.003$ & $0.575 \pm 0.003$ \\
\hline
Weighted Random & $0.586 \pm 0.002$ & $0.528 \pm 0.003$ & $0.579 \pm 0.002$ \\
\hline
\hline
Token Embedding & $0.628 \pm 0.003$ & $0.584 \pm 0.003$ & $0.622 \pm 0.003$ \\
\hline
Sentence Embedding & $0.618 \pm 0.002$ & $0.562 \pm 0.003$ & $0.610 \pm 0.002$ \\
\hline
Token MLM Loss & $0.618 \pm 0.002$ & $0.567 \pm 0.003$ & $0.607 \pm 0.002$ \\
\hline
\end{tabular}
\end{center}
\end{table}

\vspace*{-\baselineskip}
\begin{table}[ht]
\begin{center}
\caption{Results for 2017 Country Hashtag Prediction.}
\label{table:country-hashtag-prediction-2017}
\small
\begin{tabular}{|c|c|c|c|}
\hline
\textbf{\hfil{Model}} & \textbf{\hfil{Micro-F1}} & \textbf{\hfil{Macro-F1}} & \textbf{\hfil{Accuracy}} \\
\hline
Base Model 2016 & $0.418 \pm 0.003$ & $0.265 \pm 0.002$ & $0.411 \pm 0.003$ \\
\hline
Base Model 2017 & $0.561 \pm 0.003$ & $0.493 \pm 0.003$ & $0.550 \pm 0.003$ \\
\hline
\hline
Uniform Random & $0.656 \pm 0.002$ & $0.583 \pm 0.003$ & $0.646 \pm 0.002$ \\
\hline
Weighted Random & $0.656 \pm 0.002$ & $0.585 \pm 0.003$ & $0.648 \pm 0.002$ \\
\hline
\hline
Token Embedding & $0.670 \pm 0.003$ & $0.598 \pm 0.003$ & $0.660 \pm 0.003$ \\
\hline
Sentence Embedding & $0.670 \pm 0.003$ & $0.600 \pm 0.003$ & $0.660 \pm 0.003$ \\
\hline
Token MLM Loss & $0.670 \pm 0.003$ & $0.598 \pm 0.003$ & $0.661 \pm 0.003$ \\
\hline
\end{tabular}
\end{center}
\end{table}

\vspace*{-\baselineskip}
\begin{table}[ht]
\begin{center}
\caption{Results for OffensEval 2019.}
\label{table:offenseval-2019}
\small
\begin{tabular}{|c|c|c|}
\hline
\textbf{\hfil{Model}} & \textbf{\hfil{F1}} & \textbf{\hfil{AUC-ROC}} \\
\hline
Base Model 2018 & $0.515 \pm 0.013$ & $0.623 \pm 0.013$ \\
\hline
Base Model 2019 & $0.506 \pm 0.010$ & $0.636 \pm 0.011$ \\
\hline
\hline
Uniform Random & $0.606 \pm 0.019$ & $0.772 \pm 0.014$ \\
\hline
Weighted Random & $0.611 \pm 0.017$ & $0.783 \pm 0.015$ \\
\hline
\hline
Token Embedding & $0.614 \pm 0.013$ & $0.790 \pm 0.013$ \\
\hline
Sentence Embedding & $0.619 \pm 0.017$ & $0.783 \pm 0.013$ \\
\hline
Token MLM Loss & $0.618 \pm 0.016$ & $0.777 \pm 0.013$ \\
\hline
\end{tabular}
\end{center}
\end{table}

\vspace*{-\baselineskip}
\subsection{Results and Analysis}
\label{sec:result-analysis}

Table ~\ref{table:country-hashtag-prediction-2014} and ~\ref{table:country-hashtag-prediction-2017} show the results for the 2014 and 2017 Country Hashtag Prediction task, respectively. Table ~\ref{table:offenseval-2019} details the results for OffensEval 2019 task.
In all the three tables, the first two rows are the results of base models pre-trained on the tweets from the previous year and the task year, respectively. All the other rows in the tables contain the results of incremental models, which all use previous year's model as the base and incrementally train the base model with the task year's data.  

In the three tables, the results for all models follow the same trend, and we, thus, focus on the 2014 Country Hashtag Prediction task (Table ~\ref{table:country-hashtag-prediction-2014}) in the following discussion.
Our major findings are: 
\begin{itemize}
    \item On comparing ``Base Model 2013'' and ``Base Model 2014'',
    it is clear that a model trained in the past performs poorly on the new year's data, while adapting a model to new data could greatly boost its performance. This validates the necessity to keep the model informed of the evolving content.
    \item All incremental methods significantly outperform the base models.
    For instance, our proposed ``Token Embedding'' sampling method performs better than the ``Base Model 2014'' by an absolute value of 0.161 (34.5\% relatively) for Micro-F1 and 0.277 (63.6\% relatively) for Macro-F1, respectively.
     
    This suggests that the knowledge inherited from the past year (2013 base model) is still very useful, though the incremental models keep adapting to the evolving content. We claim that the incremental models for 2014 outperform the ``Base Model 2014'' as incremental models see more data: 45M examples from base model training + 24M examples from incremental training, whereas the base model sees only 45M unique examples in total.
    \item Among the incremental models, our three proposed sampling methods, ``Token Embedding'', ``Sentence Embedding'', and ``Token MLM Loss'' outperform the two baseline sampling methods by a large margin. For instance, ``Token Embedding'' performs better than the ``Weighted Sampling'' (the stronger baseline) by relatively 5.4\% and 10.6\% in Micro-F1 and  Macro-F1, respectively.
    This demonstrates the effectiveness of our proposed incremental training sampling methods. 
\end{itemize}

Note that the performance of ``Base Model 2013'' on 2014 test data is very poor in comparison to ``Base Model 2016'' on 2017 test data. We claim that this results from the larger vocabulary shift for 2014 from 2013 compared to 2017 from 2016 as seen in Figures ~\ref{fig:vocabulary_shift} and ~\ref{fig:vocabulary_shift_analysis}. Difference in the shift between 2014 from 2013 vocabulary and 2017 from 2016 vocabulary is +5.73\%, +3.44\%, and +4.83\% for natural words, wordpieces, and hashtags, respectively.

The result trends in Table ~\ref{table:country-hashtag-prediction-2017} and ~\ref{table:offenseval-2019} are very similar. The only caveat is that for the OffensEval 2019 task, the performances of ``Base Model 2018'' and ``Base Model 2019'' are comparable. This may indicate that the semantic shift for offense related language is not significant from the year 2018 to 2019. But the advantages of incremental training, and the three new incremental training sampling methods proposed by this paper are still apparent. Our proposed methods show some gains compared to baseline methods for this task, though they are not statistically significant. All these results demonstrate that our proposed sampling methods are effective.
\section{Discussion and Future Work}
\label{sec:further-discussion}

In the experiments, our three proposed sampling approaches achieve comparable results. One natural question arises, which sampling method is more suitable for real-world applications? We recommend to adopt the ``Token MLM Loss'' sampling method, where we leverage the last layer of the pre-trained BERT, mask out some tokens and then, predict the masked tokens (detailed in Section ~\ref{sec:token-mlm-loss}). This computation can be easily plugged into model online serving, and thus, perform real-time monitoring of continuously evolving content. When the overall MLM loss has an obvious increase, the system could automatically initiate incremental training process. This is more flexible and 
timely than updating the model at fixed time intervals. As a future work, we will explore the benefits of automatic incremental training and investigate our model performance during longer periods of time and other types of evolving news and social media content.

Modeling hashtags properly is important to the language model quality. In our experiments, we show that keeping popular hashtags as intact tokens in the vocabulary is very beneficial for hashtag sensitive tasks. However, there is still a large number of less popular hashtags being regarded as regular words, and thus, segmented into wordpiece tokens. In future work, we plan to explore alternative approaches for preserving hashtag information in dynamic language models.
\section{Conclusion}
\label{sec:conclusion}

In this paper, we first demonstrate the importance of dynamic modeling for continuously evolving content. Then, starting from the possibility of employing a dynamic vocabulary, we propose a simple yet effective algorithm to tackle the problem of OOV new tokens and sub-optimal tokenization. Finally, we propose three effective sampling methods to detect the training examples which contain updated knowledge and use these examples to enable efficient incremental training. We conduct extensive experiments based on two classification tasks, and demonstrate the importance of using timely content when training BERT models. We also show that our proposed sampling methods for hard example mining are not only superior to random sampling, but are also suitable for continuous model adaptation while serving live traffic.


\bibliographystyle{ACM-Reference-Format}
\bibliography{kdd}

\newpage
\begin{appendices}
\section{APPENDIX}
\label{sec:appendix}

\begin{subappendices}

\subsection{Vocabulary Shift Analysis}
\label{subsec:vocab_shift_analysis}
\begin{figure}[th]
    \centering
     \subfloat[Wordpieces]{\includegraphics[width=0.3\textwidth]{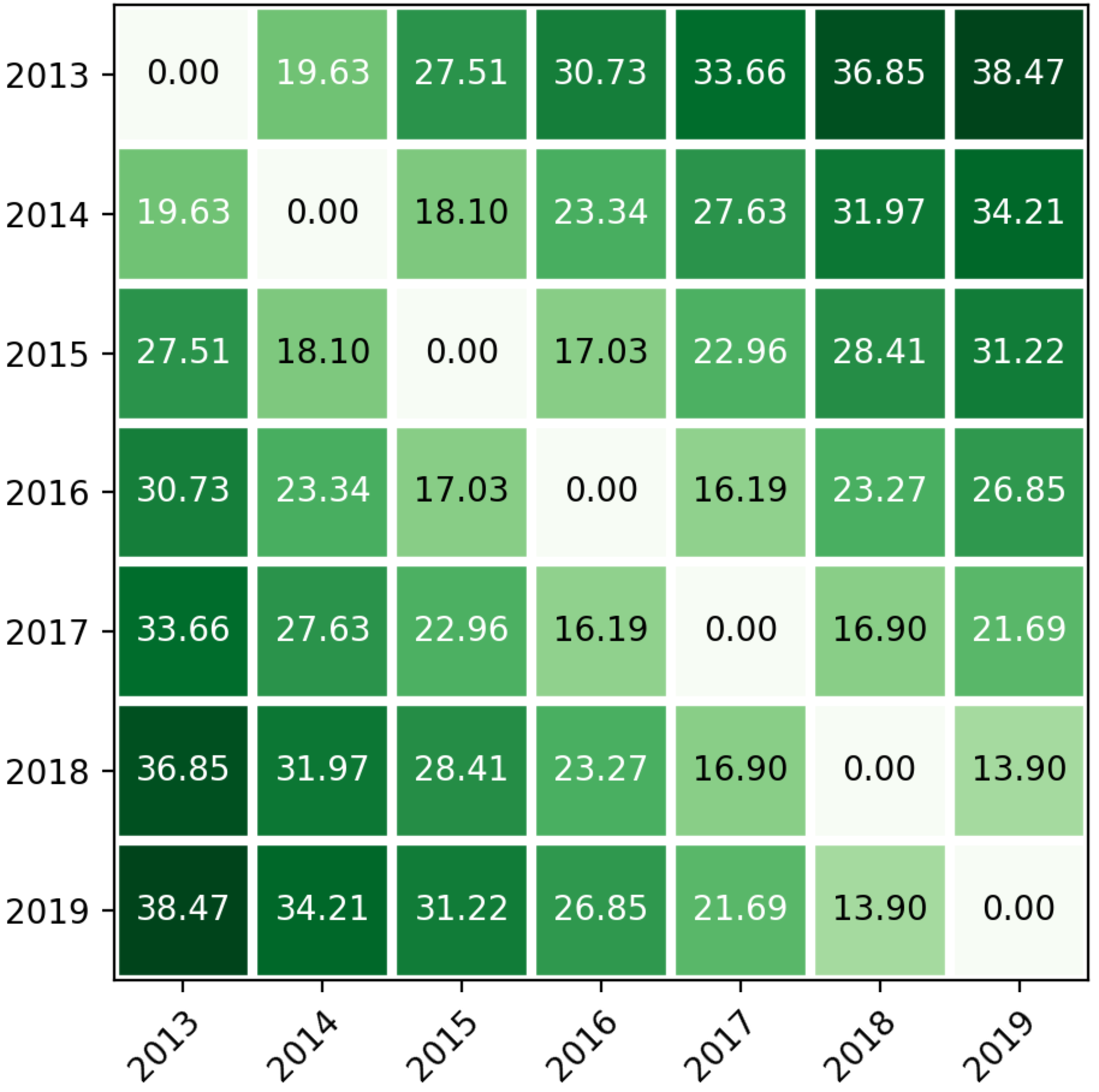}} \\
     \subfloat[Hashtags]{\includegraphics[width=0.3\textwidth]{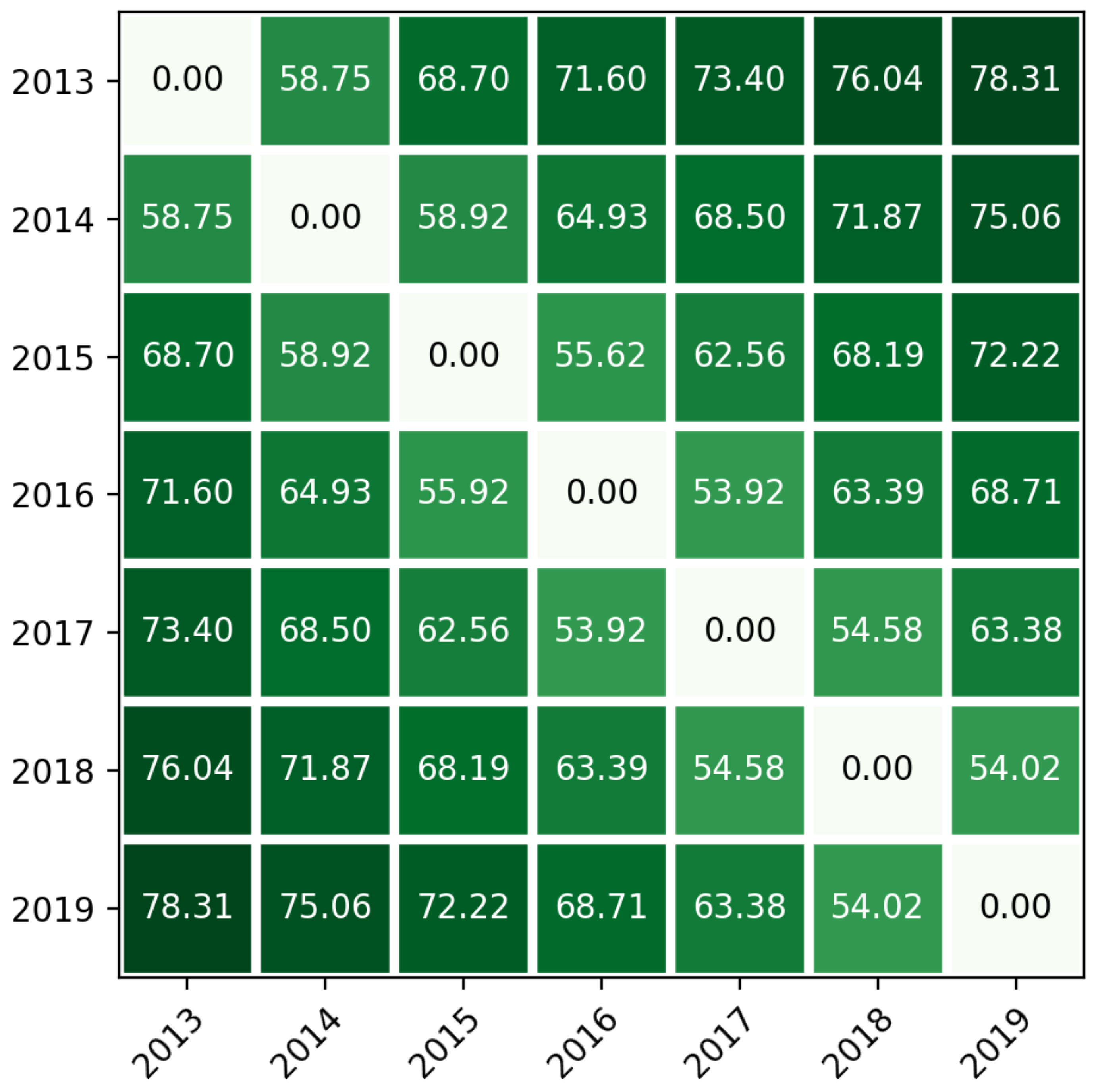}}
    \caption{Vocabulary shift (\%) for wordpieces (regular vocabulary, no hashtags) and hashtags using the top 40k tokens for the respective categories.}
    \label{fig:vocabulary_shift_analysis}
\end{figure}

In this section, we plot the vocabulary shifts between consecutive years for wordpieces and hashtags for the years 2013 -- 2019 using the top 40k tokens in each category in Figure~\ref{fig:vocabulary_shift_analysis}.

\subsection{Model Vocabulary Update}
\label{subsec:vocab_update_algo}
In Algorithm~\ref{alg:model_vocab_update}, we outline the steps for updating the model vocabulary for training, which is applicable to both base and incremental models.

\begin{algorithm}
\small
\caption{Model Vocabulary Update}
\label{alg:model_vocab_update}
\SetAlgoLined
\KwResult{Updated Model Vocabulary }
 \If{Model performance deteriorates and model needs update}{
  NewVocabulary = $\emptyset$;\\
  Fetch recent data;\\
  (Tokens, TokenCounts) = WhitespaceTokenizeRegularVocabulary(Data);\\
  (NewWordpieces, NewWordpieceCounts) = WordpieceTokenize(Tokens, TokenCounts);\\
  SortedNewWordpieces = DescendingSort(NewWordpieces, NewWordpieceCounts);\\
  NewVocabulary = CurrentVocabulary\{``wordpieces''\} $\bigcap$ NewWordpieces;\\
  SortedNewWordpieces = SortedNewWordpieces $\setminus$ NewVocabulary;\\
  \For {i = 1; i <= Count(CurrentVocabulary\{``wordpieces''\} $\setminus$ NewWordpieces); i = i + 1} {
  NewVocabulary = NewVocabulary $\bigcup$ SortedNewWordpieces[i];
  }
  \If{Vocabulary contains whole hashtags}{
    (NewHashtags, NewHashtagCounts) = WhitespaceTokenizeHashtags(Data);\\
    SortedNewHashtags = DescendingSort(NewHashtags, NewHashtagCounts);\\
    NewVocabulary = NewVocabulary $\bigcup$ (CurrentVocabulary\{``hashtags''\} $\bigcap$ NewHashtags);\\
    SortedNewHashtags = SortedNewHashtags $\setminus$ NewVocabulary;\\
    \For {i = 1; i <= Count(CurrentVocabulary\{``hashtags''\} $\setminus$ NewHashtags); i = i + 1} {
    NewVocabulary = NewVocabulary $\bigcup$ SortedNewHashtags[i];
    }
  }
  CurrentVocabulary = NewVocabulary;\\
 }
\end{algorithm}

\vspace*{-3pt}
\subsection{Data and Pre-processing}
\label{subsec:data_preprocessing}
For BERT pre-training, we use the public Twitter crawl data for the years 2013 -- 2019 available on the Internet Archive\footnote{\url{https://archive.org/details/twitterstream}}.
As a pre-processing step, we lowercase the text and replace URLs, user mentions, and emails with the special tokens ``URL'', ``@USER'', and ``EMAIL'', respectively. As each year has varied number of tweets, we randomly sample 50M unique tweets from every year for a fair comparison. These tweets are used for our initial analysis (Section~\ref{sec:dynamic_modeling}), wordpiece vocabulary generation, and BERT model pre-training (Section~\ref{sec:experiment}).

\vspace{-4pt}
\subsection{Hyperparameters}
\label{subsec:hyperparameters}
Following the original BERT paper~\cite{Devlin2019}, we mask out 15\% of tokens for pre-training which uses the MLM loss objective. 
As tweets are generally short (historically, up to 140 characters; limit has been increased to 280 characters since late 2017), we set the maximum sequence length to be 32 wordpiece tokens and mask out a max of 5 tokens per tweet. For pre-training, we use a batch size of 256 and a learning rate of 1.5e-4. For fine-tuning, we use a batch size of 32 and a learning rate of 5e-8. For other hyperparameters, we use the same values as used for training the standard BERT model~\cite{Devlin2019}.

\subsection{Effective Sampling for Incremental Training}
\label{subsec:incremental_training_algo}

Algorithm ~\ref{alg:weighted_random_sampling} details the steps for sampling hard examples for incremental training. This applies to all three sampling methods described in Section ~\ref{sec:effective_sampling_for_incremental_training}.

We perform weighted random sampling where weights are determined by a linear combination of the signal under consideration ({\it eg,} MLM loss for ``Token MLM Loss'' method) and normalized tweet length in conjunction with a random component. Here, length of a tweet is determined by the number of wordpiece tokens.

Final weights for sampling is computed as follows: \\
\begin{center}
${u^{1/({\alpha}w_{s} + (1 - {\alpha})w_{t})}}$
\end{center}
where
$u$ is a random number drawn from the uniform distribution $U (0,1)$,
$w_{s}$ is the weight from the signal (dependent on the sampling strategy),
$w_{t}$is the normalized tweet length (1.0 if tweet\_length >= 10, else $\frac{tweet\_length}{10}$), and 
$\alpha$ is the parameter controlling the contribution between the weight derived from the signal and normalized tweet length (we set it to $0.5$ in our experiments).

\vspace*{-3pt}
\begin{algorithm}
\small
\caption{Effective Sampling for Incremental Training}
\label{alg:weighted_random_sampling}
\SetAlgoLined
\KwResult{A model incrementally trained for n iterations} {
    PrecedingModel = null;\\
    CurrentModel = BaseModel;\\
    SelectedSignal = One of TokenEmbeddingShift, SentenceEmbeddingShift or TokenMLMLoss;\\
    \For{k = 1; k <= n; k = k + 1}{
        /* Assign MinWeight to m examples to sample from. */ \\
        SamplingWeights = [MinWeight] * m;\\
        \If{k = 1}{
            /* Note: For ``SentenceEmbeddingShift'', for 1st iteration, we just weight tweets by their length. */\\
            \If{SelectedSignal == TokenEmbeddingShift}{
                SamplingWeights.AdjustBy(cumulative weight of new tokens);
            }
            \If{SelectedSignal == TokenMLMLoss}{
                SamplingWeights.AdjustBy(MLM loss);
            }
        }
        \Else{
            SignalValues = SelectedSignal.Shift(CurrentModel, PrecedingModel);\\
            SamplingWeights.AdjustBy(SignalValues);\\ 
        }
        SamplingWeights.AdjustBy(tweet length);\\
        NewExamples = \textbf{WeightedRandomSample}(SamplingWeights);\\
        TrainingExamples = TrainingExamples $\bigcup$ NewExamples;\\
        NewModel = Train(CurrentModel, TrainingExamples);\\
        PrecedingModel = CurrentModel;\\
        CurrentModel = NewModel;\\
    }}
\end{algorithm}

\vspace*{-\baselineskip}
\subsection{Topics Associated With Country Hashtags}
\label{subsec:topics-with-different-country-hashtags}
In Table~\ref{table:topics-with-different-country-hashtags}, we list five of the top topics/events associated with different country hashtags.
\begin{table}[ht]
\begin{center}
\caption{Top Example Topics For Country Hashtags.}
\label{table:topics-with-different-country-hashtags}
\small
\begin{tabular}{|c|p{0.36\linewidth}|p{0.36\linewidth}|}
\hline
\textbf{\hfil{Hashtag}} & \textbf{\hfil{2014}} & \textbf{\hfil{2017}} \\
\hline
\#china & \text\hfil{\#alibaba, \#mh370, \#xinjiang, \#dalailama, obama} &  \text\hfil{\#ai, \#dangal, \#lithium, \#hres401, trump} \\
\hline
\#uk & \text\hfil{\#gaza, \#groningen, scotland, obama, go2uk} &  \text\hfil{\#ge2017, \#brexit, \#bristol, trump, ukbizz} \\
\hline
\#usa & \text\hfil{\#worldcup, \#obama, \#ibelievewewillwin, ronaldo, ebola} & \text\hfil{\#bama2017, \#trump, \#maga, \#theresistance, healthcare} \\
\hline
\end{tabular}
\end{center}
\end{table}

\subsection{Country Hashtag Prediction Task}
\label{subsec:country_hashtag_prediction_task}
For the Country Hashtag Prediction task, we collect 16 popular country hashtags 
(\#australia, \#canada, \#china, \#india, \#iran, \#iraq, \#israel, \#italy,  \#japan, \#nigeria, \#pakistan, \#philippines, \#russia, \#syria, \#uk, \#usa)
from our Twitter corpus, along with their associated tweets.
Table ~\ref{table:country-hashtag-examples} shows a few representative tweets for three of them. We use the tweets from two years, 2014 and 2017 to construct two datasets, which result in 472K tweets and 407K tweets, respectively. We remove all instances of the country hashtags and respective country names from the tweets, and randomly split them into 70\% as training, 15\% as dev, and the rest 15\% as test sets.

\begin{table}[ht]
\caption{Example Tweets associated with Country Hashtags.}
\label{table:country-hashtag-examples}
\small
\centering
\subfloat[2014\label{tab:example_tweets_2014}]{
\begin{tabular}{|c|p{0.77\linewidth}|}
\hline
\textbf{\hfil{Hashtag}} & \textbf{\hfil{Tweets}} \\
\hline
\#canada & \text\hfil{British Columbia News-  Canada launches pilot program for spouses waiting for permanent residency..  \#canada}\\
\hline
\#iran & \text\hfil{\#MaryamRajavi's Biography:The \#Iran of Tomorrow \#Women \#Lebanon \#CampLiberty \#HumanRights}\\
\hline
\#usa & \text\hfil{Aaaaand it went to the shootout but TJ Oshie wins it for \#USA over Russia! What. A. Game. \#Sochi2014}\\
\hline
\end{tabular}
}

\subfloat[2017\label{tab:example_tweets_2017}]{
\begin{tabular}{|c|p{0.77\linewidth}|}
\hline
\textbf{\hfil{Hashtag}} & \textbf{\hfil{Tweets}} \\
\hline
\#canada & \text\hfil{\#NegativeRates could hit \#Canada sooner than most expect due to \#economy’s ties to the Housing Market}\\
\hline
\#iran & \text\hfil{Guardian Council Spokesman Abbas-Ali Kadkhodaei said \#women could become candidates in the upcoming presidential elections in \#Iran.}\\
\hline
\#usa & \text\hfil{Fans flock to get new Chiefs gear after team captures AFC West title \#USA}\\
\hline
\end{tabular}
}
\centering
\end{table}

\vspace*{-\baselineskip}
\subsection{OffensEval Task}
\label{subsec:offenseval_task}
For our experiments, we use the OffensEval 2019 dataset~\cite{Zampieri2019} which contains 14K tweets posted in 2019. The original dataset has a very small test set (860 tweets). In order to have a sizable test set, we move 2240 tweets (chosen randomly) from the original training set to the test set. We further split the remainder of the original training set into training and dev sets. Our final dataset follows the ratio of 8/3/3 for train/dev/test. Table ~\ref{table:offenseval-examples} shows a few representative tweets for this task.

\begin{table}[ht]
\small
\caption{Example Tweets for OffensEval 2019.}
\label{table:offenseval-examples}
\begin{tabular}{|p{1.5cm}|p{6.5cm}|}
\hline
\textbf{\hfil{Class}} & \textbf{\hfil{Tweets}} \\
\hline
\text\hfil{OFFENSIVE} & \text\hfil{You are a fool. Denying ones free speech is deny all of our free speech.}\\
\hline
\text\hfil{NOT OFFENSIVE} & \text\hfil{Tell me did restoring your computer to an earlier date correct your problem you were having ?} \\
\hline
\end{tabular}
\end{table}

\end{subappendices}
\end{appendices}
\end{document}